%% file: main.tex
\documentclass[conference]{IEEEtran}
\usepackage{cite}
\usepackage{amsmath,amssymb,amsfonts}
\usepackage{algorithm}
\usepackage{algorithmic}
\providecommand{\Require}{\REQUIRE}

\providecommand{\State}{\STATE}
\providecommand{\Comment}[1]{\COMMENT{#1}}
\providecommand{\ForAll}{\FORALL}
\providecommand{\EndFor}{\ENDFOR}
\providecommand{\If}{\IF}
\providecommand{\Else}{\ELSE}
\providecommand{\EndIf}{\ENDIF}
\providecommand{\Return}{\RETURN}
\usepackage{graphicx}
\usepackage{textcomp}
\usepackage{xcolor}
\usepackage{booktabs}
\usepackage{multirow}
\usepackage{array}
\usepackage{tabularx}
\usepackage{url}
\usepackage[hidelinks]{hyperref}
\graphicspath{{figures/}}
\newcolumntype{L}{>{\raggedright\arraybackslash}X}
\newcolumntype{P}[1]{>{\raggedright\arraybackslash}p{#1}}
\def\BibTeX{{\rm B\kern-.05em{\sc i\kern-.025em b}\kern-.08em
    T\kern-.1667em\lower.7ex\hbox{E}\kern-.125emX}}
\begin{document}

%MARK: Title
\title{Weather Adaptive Autonomous Driving\\
}

\author{\IEEEauthorblockN{Nicholas Taylor}
\IEEEauthorblockA{\textit{Dept. of Computing and }\\
\textit{Software Engineering} \\
\textit{Florida Gulf Coast University}\\
Fort Myers, FL, USA \\
ngtaylor7853@eagle.fgcu.edu}
\and
\IEEEauthorblockN{Alexander Picón}
\IEEEauthorblockA{\textit{Dept. of Computing and }\\
\textit{Software Engineering} \\
\textit{Florida Gulf Coast University}\\
Fort Myers, FL, USA \\
aspicon5797@eagle.fgcu.edu}
\and
\IEEEauthorblockN{Sudip Dhakal}
\IEEEauthorblockA{\textit{Dept. of Computing and }\\
\textit{Software Engineering} \\
\textit{Florida Gulf Coast University}\\
Fort Myers, FL, USA \\
sdhakal@fgcu.edu}
}

\maketitle

% MARK: Abstract
\begin{abstract}
Camera object detectors lose recall in rain, fog, snow and sand, and the confidence threshold at which a detector
is deployed decides how many objects it misses and how many false alarms it raises. We study how a frozen detector
should make this decision in adverse weather. We first reproduce SeqCRC, a conformal object detector whose first
step calibrates the confidence threshold by conformal risk control (CRC), and use that step, calibrated separately
in each weather condition, as our baseline. With YOLOv8x it keeps the held-out miss fraction within a 20\% budget in every DAWN
weather bin, where a single global threshold fails in fog, but the false alarms it needs are an output rather than
a choice: from 1.6 to 117 per image across DAWN and BDD100K conditions. We then propose a calibrated trade-off
dial. One setting, $\lambda$, the number of false alarms accepted to avoid one missed object, selects the
expected-cost-optimal decision rule: either one score threshold, or a per-detection rule that keeps a box when its
calibrated probability of being a real object, estimated by gradient-boosted trees from the detector's own per-box
outputs, is at least $1/(1+\lambda)$. F-score preferences map onto the same dial, and CRC applied to $\lambda$
itself is planned to restore a per-condition miss guarantee. The dial is fitted once on BDD100K train and graded
only on BDD100K val and DAWN.
\end{abstract}


% MARK: Sections
\input{sections/introduction}
\input{sections/related_work}          % Phase 2
\input{sections/methodology}           % Phase 3
\input{sections/experimental_setup}    % Phase 1

\appendices
\input{sections/appendix_reproduction} % Phase 1 issues, fixes and evidence

\bibliographystyle{IEEEtran}
\bibliography{refs}

\end{document}

%% file: sections/introduction.tex
% MARK: Introduction (draft; the roadmap finalises it in Phase 6)
\section{Introduction}\label{sec:intro}
Autonomous vehicles must detect objects reliably when rain, fog, snow, sand, glare, and low light alter the camera
image. A detector can lose recall under these conditions even when its architecture and confidence scores remain
unchanged. Lowering a fixed confidence threshold may recover detections, but it can also produce an unacceptable
number of false alarms. The two errors do not cost the same: a false alarm may cost an unnecessary slowdown,
whereas a missed pedestrian or vehicle can cost a collision, and a miss leaves no trace at run time.

Most adverse-weather detection research retrains the detector and reports AP or mAP, which integrate over score
thresholds and leave the deployed operating point
unstated~\cite{sakaridis2018foggy,chen2018dafaster,li2023dadetect,liu2022iayolo}. Conformal methods bound what a
detector misses at its operating point~\cite{angelopoulos2024crc,andeol2025seqcrc}, but they assume that
calibration and test images are exchangeable, which persistent weather breaks, and outside
medicine~\cite{hulsman2025nodule} they do not report the false positives that a recall guarantee costs.

This work asks whether a frozen camera detector can be wrapped with a decision rule that adapts this trade-off
while providing a finite-sample miss-rate control. We first use a per-weather conformal risk control (CRC)
threshold, the first step of SeqCRC~\cite{andeol2025seqcrc} calibrated separately in each weather condition, as a
baseline. It meets its miss budget in every weather bin, but its price ranges from 1.6 to 117 false alarms per
image, and nothing lets a user choose that price. We therefore propose a calibrated per-detection trade-off dial,
in which the price is the input. So far, this work makes the following contributions:
\begin{itemize}
  \item \textbf{Reproduction.} We reproduce SeqCRC with its official code on COCO with YOLOv8x and trace its main
  discrepancy, localization sets smaller than reported, most likely to how the official YOLO wrapper rescales boxes
  when images are batched (Section~\ref{sec:repro}).
  \item \textbf{Baseline and its price.} Calibrated per weather bin, the CRC threshold holds a 20\% miss budget in
  every DAWN weather bin, where one global threshold fails in fog, and restores the budget in every BDD100K night
  bin, at 1.6 to 117 false alarms per image (Section~\ref{sec:method-baseline}).
  \item \textbf{A calibrated trade-off dial.} A single setting $\lambda$, the false alarms accepted per avoided miss,
  selects the expected-cost-optimal rule: one threshold, or a per-detection cut on a calibrated probability
  estimated from the frozen YOLOv8x's own per-box outputs. $F_\beta$ weights and a CRC miss budget map onto the same
  dial (Sections~\ref{sec:method-idea}--\ref{sec:method-design}).
  \item \textbf{An evaluation plan} on BDD100K val and DAWN, which are never used for fitting, with baselines,
  ablations and a compute budget (Section~\ref{sec:method-plan}).
\end{itemize}

Section~\ref{sec:related} reviews related work and states the research gap. Section~\ref{sec:method} describes
the baseline, the proposed dial, its architecture, algorithms and design choices, and the experiment plan.
Section~\ref{sec:repro} gives the experimental setup and the reproduction of the baseline, and
Appendix~\ref{app:repro} lists the reproduction issues and their fixes.

%% file: sections/related_work.tex
% Phase 2: Related Work (Section II of the paper). Standalone build: phase2_related_work.tex.
% MARK: Related Work
\section{Related Work}
\label{sec:related}

Our project asks whether a confidence threshold chosen per weather condition can recover the recall that a fixed
camera detector loses in adverse weather, with a finite-sample bound on the expected fraction of missed objects and a
stated false-positive cost. Four lines of work bear on this question: methods and benchmarks that make detection robust
to weather, studies of how detectors' confidences and operating thresholds behave under shift, distribution-free
guarantees for detection, and risk control under shift or within groups. Table~\ref{tab:related} lists the method,
input, data, headline result and main limitation of each paper we discuss.

\begin{table*}[!t]
\centering
\caption{Related work at a glance (one row per paper; AP and mAP as defined in each paper).}
\label{tab:related}
\scriptsize
\setlength{\tabcolsep}{3pt}
\renewcommand{\arraystretch}{1.12}
\begin{tabularx}{\textwidth}{@{}P{2.05cm}P{3.2cm}P{1.9cm}P{2.6cm}LL@{}}
\toprule
\textbf{Paper} & \textbf{Method} & \textbf{Sensor / input} & \textbf{Dataset} & \textbf{Key result} & \textbf{Limitation} \\
\midrule
\multicolumn{6}{@{}l}{\textit{Weather-robust detection and benchmarks}}\\
Sakaridis~\cite{sakaridis2018foggy}, IJCV'18 & Physics-based synthetic fog; fog fine-tuning & RGB camera & Foggy Cityscapes; Foggy Driving (101 real) & Car AP 30.5$\to$35.3 on 33 finely annotated real images & Homogeneous daytime fog; tiny real test set \\
Hassaballah~\cite{hassaballah2021dawn}, T-ITS'21 & Visibility enhancement + multi-scale CNN + GM-PHD tracker; introduces DAWN & RGB camera & DAWN (fog, rain, snow, sand), KITTI, COCO & Reported to outperform 21 compared vehicle detectors & Small real-image set (1,027 released images); no official splits \\
Chen~\cite{chen2018dafaster}, CVPR'18 & Adversarial image + instance alignment (DA Faster R-CNN) & RGB camera & Cityscapes$\to$Foggy Cityscapes; SIM10k; KITTI & Foggy mAP@0.5 18.8$\to$27.6 & Retrains per shift; synthetic fog; AP/mAP only, no operating point \\
Li~\cite{li2023dadetect}, WACV'23 & Hardness-weighted GRL + rainy-domain triplet loss (DA-Detect) & RGB camera & Cityscapes$\to$Foggy Cityscapes; KITTI & Foggy mAP@0.5 23.41$\to$42.34 (oracle 45.6) & Synthetic fog only; 3.5-pt hyperparameter swing; no operating point \\
Liu~\cite{liu2022iayolo}, AAAI'22 & CNN-predicted differentiable filters before YOLOv3 & RGB camera & VOC-Foggy/Dark; RTTS; ExDark & RTTS mAP 28.82$\to$37.08 (30.80 for same-data YOLOv3) & Retrains detector; mAP only \\
\midrule
\multicolumn{6}{@{}l}{\textit{Confidence calibration and operating thresholds}}\\
Oksuz~\cite{oksuz2023selfaware}, CVPR'23 & Self-aware detector: image rejection, calibration, LRP-optimal thresholds & RGB camera & nuImages$\to$BDD45K, BDD45K-C & BDD45K (F-RCNN, IoU 0.1): 1$-$recall 58.1\%, 1$-$precision 13.2\% & Val-tuned thresholds, no guarantee; weather pooled \\
Kuzucu~\cite{kuzucu2024calibration}, ECCV'24 & Calibration and accuracy on the thresholded set; Platt/isotonic baselines & RGB camera & COCO, Cityscapes, LVIS; corrupted, Foggy Cityscapes, Obj45K & Foggy Cityscapes LaECE$_0$ 18.5$\to$10.0 (isotonic) & In-distribution thresholds reused under shift; no guarantee \\
\midrule
\multicolumn{6}{@{}l}{\textit{Distribution-free guarantees for detection}}\\
Angelopoulos~\cite{angelopoulos2024crc}, ICLR'24 & CRC: scalar parameter bounding expected monotone loss & Any model's scores & Polyps, COCO, ImageNet, NQ & COCO mean FNR 0.0996 at $\alpha{=}0.1$ (1,000 splits) & Marginal only; shift needs weights or TV slack \\
And\'eol~\cite{andeol2023railway}, COPA'23 & Conformal box margins; image-wise recall CRC & RGB cab-view camera & Railway signals, 3,414 images & DiffusionDet stretch 1.56 (pixel CRC) vs.\ 9.83 (Hausdorff CP), risk 0.09 & Fixed 0.3 threshold; ignores false positives \\
Timans~\cite{timans2024adaptive}, ECCV'24 & Two-step CP: class sets, then box intervals & Detector outputs (RGB) & COCO, Cityscapes, BDD100K & COCO box coverage 0.88 (target 0.9) vs.\ 0.21 deep ensemble & Detected objects only; misses uncovered \\
Doula~\cite{doula2024arcp}, CVPRW'24 & Conformal class sets for detected agents; taxonomy parent for AR display & RGB camera video & ROAD; user study (ROAD, Seeing Through Fog, Gated2Gated) & APS set size 6.16$\to$1.00 at coverage 0.90 ($\alpha{=}0.1$) & Classes of detected agents only; coverage pooled over conditions \\
And\'eol~\cite{andeol2025seqcrc}, arXiv'25 & Sequential CRC: confidence threshold, then box/class sets & Detector outputs (RGB) & COCO val (2,500 cal./2,500 test) & YOLOv8x box-count-recall risk 0.019 ($\alpha_{\mathrm{cnf}}{=}0.02$), 11.7 boxes/img & False positives not bounded; i.i.d.\ COCO only \\
Hulsman~\cite{hulsman2025nodule}, COPA'25 & CRC on a frozen detector's score threshold; per-scan miss-rate loss & 3-D thoracic CT & LIDC-IDRI, four reader-consensus subsets & $\geq$3 readers: 91.35\% sensitivity at 2.25 FP/scan ($\alpha{=}0.1$) & Exchangeability assumed; 118 FP/scan on low-consensus set \\
\midrule
\multicolumn{6}{@{}l}{\textit{Risk control under shift and within groups}}\\
Farinhas~\cite{farinhas2024nonexch}, ICLR'24 & Weighted CRC with total-variation penalty & Any model's scores & Synthetic time series; ELEC2; NQ & FNR 0.246$\to$0.196 at $\alpha{=}0.2$ (changepoints) & TV penalty unknown; no detection \\
Blot~\cite{blot2025aacrc}, AISTATS'25 & CRC conditional on a function class (incl.\ group indicators) & Any model's scores (images) & Polyp and fire segmentation; simulated regression & Polyp recall 0.906 (target 0.9); precision 0.395$\to$0.457 & Assumes exchangeability; no detection or weather shift \\
\bottomrule
\end{tabularx}
\end{table*}

% MARK: Weather-Robust Detection and How It Is Measured
\subsection{Weather-Robust Detection and How It Is Measured}
Most adverse-weather detection research changes the detector or its
input~\cite{sakaridis2018foggy,chen2018dafaster,li2023dadetect,liu2022iayolo}. Sakaridis \textit{et
al.}~\cite{sakaridis2018foggy} render homogeneous fog onto clear Cityscapes images with the Koschmieder model, creating
Foggy Cityscapes. Domain-adaptive detectors use such synthetic fog without target labels: DA Faster R-CNN~\cite{chen2018dafaster} aligns image- and instance-level features adversarially, while Li \textit{et al.}~\cite{li2023dadetect} add hardness weighting and a triplet loss for synthetic rain. Image-adaptive YOLO~\cite{liu2022iayolo} instead predicts differentiable defogging, colour, tone, and sharpening filters jointly with
YOLOv3. These methods improve detection on their target benchmarks, but their results depend on retraining and are
usually reported as AP or mAP rather than at a specified deployment threshold. Real
adverse-weather test sets also remain relatively small: Foggy Driving has 101 real images~\cite{sakaridis2018foggy}, and
DAWN~\cite{hassaballah2021dawn} contains real traffic images in fog, rain, snow and sand. The paper that introduces DAWN
also proposes a visibility-enhancement front end, a multi-scale CNN detector and a GM-PHD tracker, compared with 21
vehicle detectors; DAWN's four subsets define our weather conditions.

The common limitation is that these methods retrain the network for a particular shift~\cite{sakaridis2018foggy,chen2018dafaster,li2023dadetect,liu2022iayolo}. Because AP and mAP integrate over score
thresholds, they do not directly state the miss rate or false-positive cost at a deployed operating point. Each method
therefore yields a detector that still needs an operating point,
which makes this line complementary to ours.

% MARK: Confidence, Calibration and Operating Thresholds
\subsection{Confidence, Calibration and Operating Thresholds}
A second line studies whether detector scores can be trusted and where to cut them. Oksuz \textit{et
al.}~\cite{oksuz2023selfaware} argue that a deployed detector must be judged at its operating threshold and build a
self-aware detector whose class-wise thresholds are tuned on in-distribution validation data. Trained on nuImages and
tested on BDD45K, their Faster R-CNN variant errs mostly by missing objects at the selected operating point.
Kuzucu \textit{et al.}~\cite{kuzucu2024calibration} show that common protocols score
calibration and accuracy on different detection sets, and evaluate both on the thresholded detections a system would
deploy. On Foggy Cityscapes, isotonic regression improves localization-aware calibration, yet fog still degrades
localization performance. Read together, the two studies show that shift hurts a detector at its operating point,
mainly through misses~\cite{oksuz2023selfaware}, and that
recalibration can repair confidence while accuracy still falls~\cite{kuzucu2024calibration}. Yet both tune class-wise
thresholds once on validation data from the training distribution and carry them to shifted test data, so neither
guarantees a miss rate in the shifted condition.

% MARK: Distribution-Free Guarantees for Detection
\subsection{Distribution-Free Guarantees for Detection}
Conformal risk control (CRC)~\cite{angelopoulos2024crc} chooses the least conservative value of a scalar parameter
whose corrected calibration risk is at most a budget $\alpha$; for exchangeable data and a bounded loss that is
non-increasing in the parameter and within budget at its most permissive value, the expected test loss is then at most
$\alpha$. Conformal methods for detection differ mainly in which quantity they calibrate, and that choice decides
whether objects the detector never finds are covered. And\'eol \textit{et al.}~\cite{andeol2023railway} calibrate box
margins for railway-signal detection at a fixed objectness threshold of 0.3, with image-wise recall losses that need far
smaller margins than a Hausdorff-coverage baseline, and state that they do not consider false positives. Timans \textit{et
al.}~\cite{timans2024adaptive} first build conformal class sets and then box intervals; their intervals approach the
target coverage on COCO but only for objects detected above a fixed score. AR-CP~\cite{doula2024arcp} is the one
conformal method we found aimed at adverse driving conditions: it
builds class prediction sets for agents detected on the ROAD dataset and replaces large sets by their common parent
class for an augmented-reality display, with coverage reported over all conditions together.
SeqCRC~\cite{andeol2025seqcrc} calibrates the confidence threshold itself and then the box and class sets, so that
undetected objects enter the guaranteed quantity, but it does not bound a false-positive quantity. Outside
driving, Hulsman \textit{et al.}~\cite{hulsman2025nodule} apply single-parameter CRC to the score threshold of a frozen
3-D lung-nodule detector with a per-scan false-negative-rate loss and report the corresponding false-positive price,
which varies substantially across annotation-consensus subsets.

Across this line, a guarantee counts missed objects only when the confidence threshold itself is calibrated with a
recall-type loss~\cite{andeol2025seqcrc,hulsman2025nodule}, or when boxes at a fixed threshold are enlarged until they
cover them, which fails on images without any detection: the railway budget is unattainable for two of its three
pretrained detectors~\cite{andeol2023railway}. Coverage of matched detections~\cite{timans2024adaptive,doula2024arcp}
ignores missed objects altogether. All of these studies assume that calibration and test images are exchangeable,
splitting the pool at most by object class~\cite{timans2024adaptive}. The closest to per-condition calibration is the
separate calibration of each annotation-consensus subset in~\cite{hulsman2025nodule}, but these subsets are nested label
definitions, not deployment conditions. That study is also the only one to report false positives at its guaranteed
recall; SeqCRC reports only how many boxes it keeps.

% MARK: Risk Control Under Shift and Within Groups
\subsection{Risk Control Under Shift and Within Groups}
Weather persists for minutes and may be rare in a calibration mix, so the test condition is not a random draw from the
pool. Under arbitrary shift, the CRC bound loosens by the loss bound $B$ times the summed total-variation distance
between calibration and test distributions~\cite{angelopoulos2024crc}. Non-exchangeable CRC~\cite{farinhas2024nonexch}
down-weights calibration points believed to be far from the test point and improves control under synthetic
changepoints, but the total-variation penalty in its bound is unknown in
practice, so the weights are heuristics. Blot \textit{et al.}~\cite{blot2025aacrc} keep exchangeability but recast CRC
as an optimization over a class of functions of the input, which yields risk control within each group when the class
is spanned by group indicators; with a learned embedding they meet a segmentation recall target at higher precision
than marginal CRC. The two routes differ in what they need: weighting
leaves an unknown total-variation term in its bound, whereas group conditioning needs only a group label for each image,
which DAWN's weather subsets supply. Neither has been applied to the confidence threshold of an object detector under
weather.

% MARK: Research Gap
\subsection{Research Gap}
Prior work leaves three gaps. First, weather-robust
detectors~\cite{sakaridis2018foggy,chen2018dafaster,li2023dadetect,liu2022iayolo} improve AP or mAP but leave the
operating point and its miss rate unstated, and threshold-selection
studies~\cite{oksuz2023selfaware,kuzucu2024calibration} tune class-wise thresholds on in-distribution data without a
guarantee. Second, conformal detection methods~\cite{andeol2023railway,timans2024adaptive,doula2024arcp,andeol2025seqcrc}
calibrate under an exchangeability assumption, at most split by object class, and, except in
medicine~\cite{hulsman2025nodule}, do not report how many false positives a recall guarantee costs; the one study in
adverse driving conditions covers only the classes of objects already detected~\cite{doula2024arcp}. Third,
group-conditional risk control has been developed~\cite{blot2025aacrc}, but to our knowledge it has not been used to
calibrate a frozen camera detector's confidence threshold per weather condition.

Our work addresses all three. We keep a COCO-pretrained YOLOv8 detector fixed. As a baseline, we apply CRC
separately in each weather condition of DAWN with an IoU- and class-matched per-image miss loss, compare these
thresholds with one pooled threshold, report the false positives per image at every budget, and flag budgets that no
threshold can certify because some objects are never detected. We then make the false-positive price an input
rather than an output: one setting $\lambda$, the number of false alarms accepted per avoided miss, selects the
expected-cost-optimal rule, either one threshold or a per-detection cut on a calibrated probability that each box is
real, and conformal risk control applied to $\lambda$ itself (planned) restores a per-condition miss guarantee.
Unlike perception-side methods, our approach needs no retraining and can wrap any of them; unlike prior conformal
detection, its guarantee is stated per weather condition and comes with its false-positive price.

%% file: sections/methodology.tex
% Phase 3: Methodology (Section III of the paper).
% MARK: Methodology
\section{Methodology}\label{sec:method}

% MARK: Baseline: per-weather conformal threshold (Phase 1)
\subsection{Baseline: per-weather conformal threshold (Phase~1)}\label{sec:method-baseline}

Our baseline is the first step of SeqCRC~\cite{andeol2025seqcrc}: conformal risk control
(CRC)~\cite{angelopoulos2024crc} on the confidence threshold of a frozen detector, with an image-level loss
$L_i(\tau)$, the fraction of ground-truth objects in calibration image~$i$ with no matched detection scoring at
least~$\tau$. Calibrated separately in each weather bin~$b$ with $n$ calibration images and mean calibration
loss $\hat R(\tau; b) = \tfrac{1}{n}\sum_{i=1}^{n} L_i(\tau)$, the threshold is
\begin{equation}
  \hat\tau(b) \;=\; \max\Big\{\tau \;:\; \tfrac{n}{n+1}\,\hat R(\tau; b) + \tfrac{1}{n+1} \le \alpha\Big\},
  \label{eq:baseline-crc}
\end{equation}
which bounds the expected miss fraction of an exchangeable test image from the same bin by~$\alpha$.
In Phase~1 we reproduced SeqCRC with its official code on COCO, where the global test risk stayed within its
budget in every configuration we ran (see Section~\ref{sec:repro} for our Phase~1 report), and ran it on
DAWN~\cite{kenk2020dawn}. Our harness implements Eq.~\eqref{eq:baseline-crc} per weather bin on
YOLOv8x~\cite{ultralytics2023yolov8}. With $\alpha = 0.2$ on DAWN, the per-bin threshold keeps the held-out miss
fraction at 0.188--0.194 in every bin, at 1.6--6.2 false positives per image (FPPI), while one global threshold
misses 0.251 in fog (above $\alpha$ in 98\% of 200 splits). On BDD100K~\cite{yu2020bdd100k}, a threshold
calibrated on clear days misses 0.28--0.36 at night; per-bin thresholds restore the budget at 43--117 FPPI.

% MARK: Limitation (the gap)
\paragraph{Limitation (the gap)} The baseline fixes the miss budget and lets the false alarms be whatever
results: at night it buys $\alpha = 0.2$ with up to 117 false alarms per image, and there is no way to ask for a
different balance. It applies one score cut to every box, it needs the weather bin at run time (our camera-only
weather estimators did not isolate fog), and its guarantee assumes exchangeable data within each bin.

% MARK: Proposed method: a calibrated trade-off dial
\subsection{Proposed method: a calibrated trade-off dial}\label{sec:method-idea}

\textbf{Core idea.} Instead of one confidence threshold per weather bin, every detection receives a calibrated
probability $p(d)$ that it is a real object, computed from the frozen detector's own outputs, and is kept if and
only if $p(d) \ge 1/(1+\lambda)$. The single user-facing parameter $\lambda$, the number of false alarms
accepted to avoid one missed object, sets the trade-off for the use case. An F-score weight $\beta$ (and, in
Phase~4, a CRC miss budget $\alpha$) maps onto the same dial.

Table~\ref{tab:changes} lists what changes relative to the baseline. The detector, data, matching rule and
evaluation protocol stay the same; only the decision of which boxes to keep is replaced.

\begin{table}[t]
\centering\footnotesize
\caption{What our method keeps, replaces and adds relative to the Phase-1 baseline.}
\label{tab:changes}
\begin{tabularx}{\linewidth}{@{}>{\raggedright\arraybackslash}p{1.6cm}>{\raggedright\arraybackslash}X>{\raggedright\arraybackslash}X@{}}
\toprule
 & Baseline (SeqCRC step~1 per weather bin) & Ours (trade-off dial) \\
\midrule
Detector & YOLOv8x, frozen COCO weights, 1280\,px, score floor 0.001 & unchanged \\
Decision rule & one score threshold per weather bin, $s \ge \hat\tau(b)$ & per detection, $p(d) \ge 1/(1+\lambda)$; simple option: one threshold $s \ge \tau(\lambda)$ \\
Trade-off & fixed miss budget $\alpha$; false alarms unpriced & explicit input $\lambda$ (or $\beta$); $\alpha$ planned as a floor via CRC over $\lambda$ \\
Run-time inputs & score and the weather bin & score, box context and the detector's per-box outputs (no weather label) \\
Fitting & calibration half of each bin & once, on 21,831 labelled BDD100K train frames \\
Added & --- & per-box feature tap; calibrated model; $\beta \to \lambda$ map; label-free estimate of false alarms \\
\bottomrule
\end{tabularx}
\end{table}

% MARK: Architecture
\subsection{Architecture}\label{sec:method-arch}

\begin{figure*}[t]
  \centering
  % Diagram v2 (Oct. 29, 2026). The first version is kept as figures/architecture.pdf (and .svg);
  % write {architecture} instead of {architecturev2} to go back to it.
  \includegraphics[width=\textwidth]{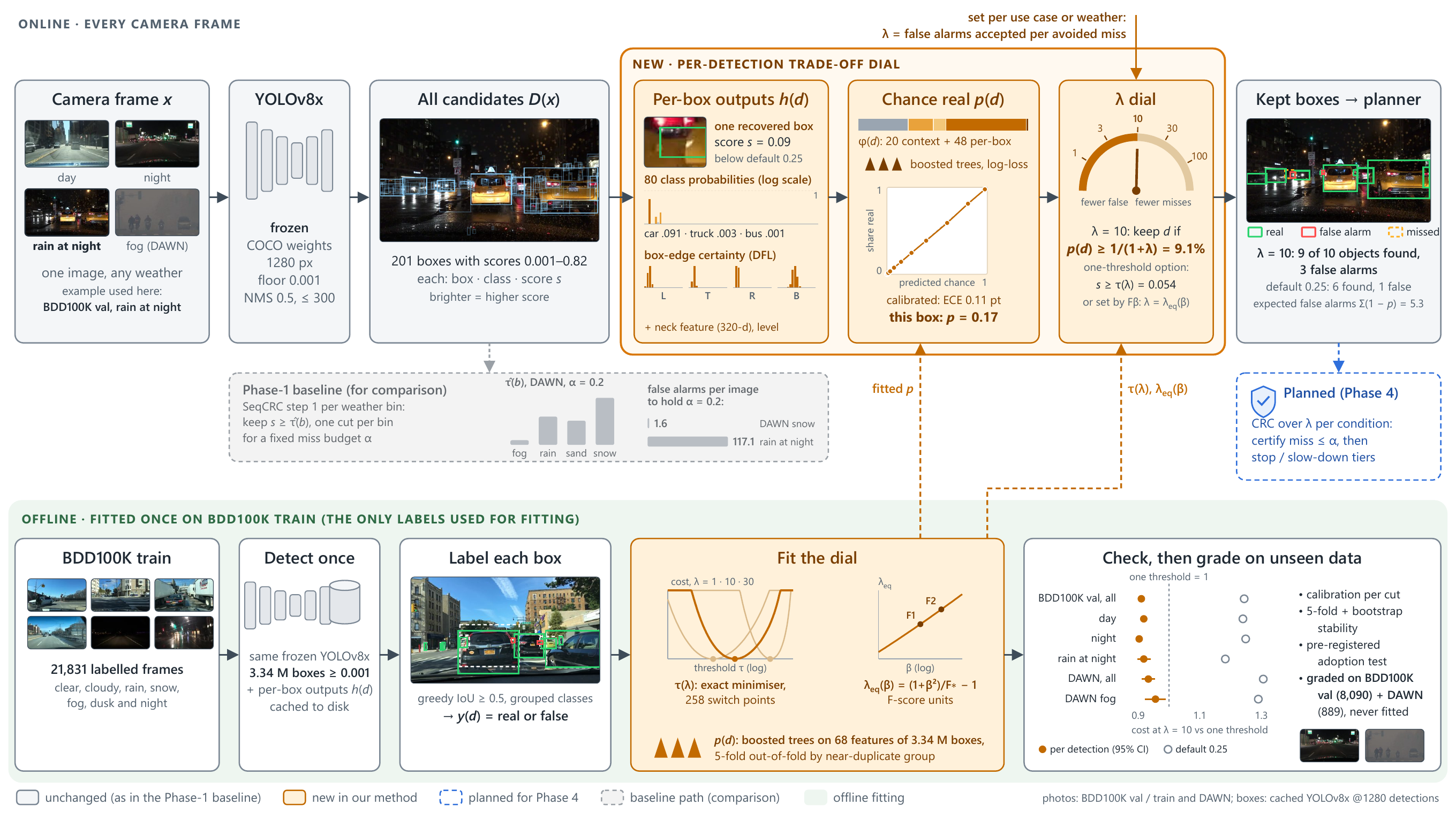}
  \caption{Architecture, following one BDD100K val frame (rain at night) through the online path. New components
  are orange, unchanged ones grey; the dashed grey lane is the Phase-1 baseline, kept for comparison, and the
  dashed blue box is planned for Phase~4. Fitting (green band) uses only BDD100K train; grading uses BDD100K val
  and DAWN.}
  \label{fig:architecture}
\end{figure*}

Figure~\ref{fig:architecture} shows the full pipeline. Online, YOLOv8x runs once per camera frame and returns
candidate boxes $D(x)$ with class and score~$s$. We also keep the detector's per-box outputs $h(d)$ at the
anchor each box came from: the 80 class probabilities, the four distribution-focal-loss (DFL) distributions over
the box sides~\cite{li2020gfl}, the neck feature vector, and the pyramid level. A feature builder forms
$\varphi(d)$: 20 context features (score, class, size and position, rank, overlap with stronger boxes, crowding)
plus 48 per-box features (six class probabilities, strongest other class, top-two margin, sum and entropy of the
probabilities, DFL entropy per side and mean, level, and 32 principal components of the neck feature).
A gradient-boosted model maps $\varphi(d)$ to $p(d)$, and the dial keeps or drops each box. Offline, the same
frozen detector is run once over BDD100K train, boxes are labelled by matching, and the dial is fitted, checked
and then graded on data never used for fitting.

% MARK: Algorithm
\subsection{Algorithm}\label{sec:method-alg}

% MARK: Cost
\paragraph{Cost} For a keep rule $k$ over the detections of image $x$, let $M_k(x)$ be the number of
ground-truth objects left without a kept matched box and $F_k(x)$ the number of kept boxes that match no object.
The dial setting $\lambda > 0$ defines the cost
\begin{equation}
  C_\lambda(k) \;=\; \sum_{x}\big[\lambda\, M_k(x) + F_k(x)\big].
  \label{eq:cost}
\end{equation}

% MARK: Option A: one threshold
\paragraph{Option A: one threshold} $k(d) = \mathbf{1}[s(d) \ge \tau(\lambda)]$ with
\begin{equation}
  \tau(\lambda) \;=\; \arg\min_{\tau} \sum_{x \in \mathcal{D}_{\text{train}}} \big[\lambda\, M_\tau(x) + F_\tau(x)\big].
  \label{eq:one-threshold}
\end{equation}
Greedy non-maximum suppression and greedy matching are both score-ordered, so filtering one detector run at
$\tau$ equals running the detector at $\tau$. Equation~\eqref{eq:one-threshold} is therefore solved exactly from
cumulative counts: the minimisers for all $\lambda$ are the corners of the lower convex hull of
(real kept, false kept), stored as 258 switch points.

% MARK: Option B: per detection
\paragraph{Option B: per detection} Keeping a box with probability $p$ of being real costs an expected
$1-p$ false alarms; dropping it costs an expected $\lambda p$ in misses. The cost-optimal rule is
therefore~\cite{elkan2001foundations}
\begin{equation}
  \text{keep } d \iff 1 - p(d) \le \lambda\, p(d) \iff p(d) \ge \frac{1}{1+\lambda},
  \label{eq:bayes}
\end{equation}
which is optimal when $p$ is calibrated. Because the kept set grows with $\lambda$, the miss count is
non-increasing in $\lambda$. We estimate $p(d) = \sigma\big(g(\varphi(d))\big)$ with gradient-boosted
trees~\cite{friedman2001gbm,ke2017lightgbm,pedregosa2011sklearn}. They are trained on all train detections by
minimising the log-loss
$\ell = -\sum_d \big[y_d \log p_d + (1-y_d)\log(1-p_d)\big]$, where $y_d \in \{0,1\}$ comes from greedy
IoU-$0.5$ matching with grouped classes (vehicle, two-wheeler, person). We check calibration with the expected
calibration error~\cite{guo2017calibration} and with the share of real boxes among detections near each cut
$1/(1+\lambda)$.

% MARK: F-score units
\paragraph{F-score units} With calibrated probabilities, the $F_\beta$-optimal cut is $F^*_\beta/(1+\beta^2)$,
where $F^*_\beta$ is the best achievable $F_\beta$~\cite{zhao2013beyond,lipton2014optimal}. Setting this cut
equal to $1/(1+\lambda)$ gives the equivalent dial setting
\begin{equation}
  \lambda_{\text{eq}}(\beta) \;=\; \frac{1+\beta^2}{F^*_\beta} - 1,
  \label{eq:fbeta}
\end{equation}
with $F^*_\beta$ measured on train for each option. For example, $F_1$ corresponds to $\lambda \approx 1.8$ and
$F_2$ to $\lambda \approx 6.2$ with the per-detection option.

% MARK: Planned: a miss budget on top of the dial
\paragraph{Planned: a miss budget on top of the dial (CRC over \texorpdfstring{$\lambda$}{lambda})} Because the image-level miss fraction
$L_i(\lambda)$ is non-increasing in $\lambda$, the baseline's rule~\eqref{eq:baseline-crc} applies to $\lambda$
itself. Per condition $c$ with $n_c$ calibration images,
\begin{equation}
  \hat\lambda_c \;=\; \inf\Big\{\lambda \;:\; \tfrac{n_c}{n_c+1}\,\hat R_c(\lambda) + \tfrac{1}{n_c+1} \le \alpha\Big\},
  \label{eq:crc-lambda}
\end{equation}
and deploying $\max(\lambda_{\text{user}}, \hat\lambda_c)$ keeps the CRC guarantee
$\mathbb{E}[L_{\text{test}}] \le \alpha$ under exchangeability within the condition, while never being less
cautious than the user asked. The fit is infeasible when $\alpha$ is below the miss fraction at the score floor
(11--18\% of objects are never detected).

\begin{algorithm}[t]
\caption{Fitting the dial (offline, BDD100K train only)}\label{alg:fit}
\begin{algorithmic}[1]
\Require labelled train frames $\mathcal{D}_{\text{train}}$; frozen detector $f$
\ForAll{$x \in \mathcal{D}_{\text{train}}$}
  \State $(D(x), h) \gets f(x)$ \Comment{floor 0.001, NMS 0.5, $\le 300$ boxes; keep per-box outputs}
  \State $y(d) \gets$ greedy IoU $\ge 0.5$ match of $d \in D(x)$ to labels \Comment{grouped classes}
\EndFor
\State \textbf{Option A:} sort scores; keep ends of equal-score blocks; cumulative real and false counts
\State \hspace{1.2em} $\tau(\cdot) \gets$ corners of the lower convex hull \Comment{all $\lambda$ at once}
\State \textbf{Option B:} $\varphi(d) \gets [\text{context}(d, D(x)),\ \text{per-box}(h(d))]$ \Comment{PCA fitted on train}
\State \hspace{1.2em} $g \gets$ gradient boosting on $\{(\varphi(d), y(d))\}$ \Comment{log-loss}
\State \hspace{1.2em} $p_{\text{oof}} \gets$ 5-fold out-of-fold probabilities, folds by near-duplicate group
\ForAll{$\beta$}
  \State $\lambda_{\text{eq}}(\beta) \gets (1+\beta^2)/F^*_\beta - 1$ \Comment{F-score map, $F^*_\beta$ on train}
\EndFor
\State check calibration near every cut $1/(1+\lambda)$ and fold stability; adopt only if the pre-registered criteria hold
\State \Return $\tau(\cdot)$, $g$ (with PCA), $\lambda_{\text{eq}}(\cdot)$
\end{algorithmic}
\end{algorithm}

\begin{algorithm}[t]
\caption{Applying the dial (online, every camera frame)}\label{alg:deploy}
\begin{algorithmic}[1]
\Require frame $x$; setting $\lambda$ (or $\beta$); option $\in\{$A, B$\}$
\If{$\beta$ is given} \State $\lambda \gets \lambda_{\text{eq}}(\beta)$ \Comment{F-score map} \EndIf
\If{a budget $\alpha$ is certified for condition $c$} \State $\lambda \gets \max(\lambda, \hat\lambda_c)$ \Comment{planned: CRC over $\lambda$} \EndIf
\State $(D, h) \gets f(x)$ \Comment{one detector pass}
\ForAll{$d \in D$}
  \If{option A}
    \State keep $d$ iff $s(d) \ge \tau(\lambda)$
  \Else
    \State $p(d) \gets \sigma(g(\varphi(d)))$; keep $d$ iff $p(d) \ge 1/(1+\lambda)$ \Comment{cost-optimal cut}
  \EndIf
\EndFor
\State $\widehat{\mathrm{FA}} \gets \sum_{d \text{ kept}} (1 - p(d))$ \Comment{expected false alarms, no labels needed}
\State \Return kept detections, $\widehat{\mathrm{FA}}$
\end{algorithmic}
\end{algorithm}

% MARK: Design choices
\subsection{Design choices}\label{sec:method-design}

Table~\ref{tab:design} gives each design choice, the reason for it, and the alternatives we tried or
considered. Most of the alternatives were run in the harness, several under pre-registered criteria.

\begin{table*}[t]
\centering\small
\caption{Design choices, the reasons for them, and the alternatives considered (with the evidence).}
\label{tab:design}
\begin{tabularx}{\textwidth}{@{}>{\raggedright\arraybackslash}p{3.4cm}>{\raggedright\arraybackslash}X>{\raggedright\arraybackslash}X@{}}
\toprule
Choice & Why & Alternatives considered \\
\midrule
Keep the detector frozen; decide after it &
calibration stays valid; one detector run serves every setting; no retraining per condition &
domain adaptation (DA-Detect: 20.41 vs 21.44 AP@50 on real DAWN fog in Phase~1); fine-tuning (invalidates every calibration) \\
Trade-off as $\lambda$, false alarms per avoided miss &
use case sets the balance; the optimal rule is explicit (Eq.~\ref{eq:bayes}); $\beta$ and $\alpha$ map onto it &
fixed $\alpha$ only (baseline, false alarms unpriced); maximise $F_1$ (one fixed balance); FPPI cap \\
Decide per detection &
87--91\% of a per-image oracle's advantage on BDD100K is chance; the learnable signal is in individual boxes &
per-image thresholds from pixels, from the confidence profile, or from YOLOv8x image features (all failed pre-registered tests); box-combination rules; summed-confidence maps \\
One threshold kept as the simple option &
one auditable number per $\lambda$; per-weather thresholds give no gain (0.98--1.03$\times$ the cost) &
per-weather thresholds (the baseline's form) \\
Gradient-boosted trees on detector outputs &
tabular inputs; trains in minutes on CPU; calibrated without post-hoc scaling (expected calibration error 0.11 points on BDD100K val) &
recalibrating the score alone (this is Option~A); a neural head; retraining YOLOv8x's head \\
YOLOv8x's own per-box outputs as features &
pre-registered gain: cost 0.972$\times$ (0.969--0.975) of the context-only model on BDD100K, not worse on DAWN (0.995) &
context features only (model~1); a separate CNN on box crops (more compute) \\
One model for every $\beta$ &
theory: only the cut moves with $\beta$; dedicated $\beta$-weighted models tie (0.997--1.001$\times$) &
one cost-weighted model per $\beta$ \\
Fit on BDD100K train; grade on BDD100K val and DAWN &
nothing is chosen on graded data; DAWN tests transfer to real fog, rain, sand and snow &
calibrating on halves of the test pool (the baseline protocol) \\
Grouped classes in matching &
car/truck confusions are not safety misses &
class-aware matching (run~1) \\
\bottomrule
\end{tabularx}
\end{table*}

% MARK: Experiment plan (Phase 4)
\subsection{Experiment plan (Phase~4)}\label{sec:method-plan}

% MARK: Data and splits
\paragraph{Data and splits} The dial is fitted on 21,831 BDD100K train frames (3.34\,M detections), with 5-fold
out-of-fold estimates grouped by near-duplicate group. It is graded on BDD100K val (8,090 frames in seven
weather-by-time bins) and on DAWN (889 frames with clean labels: fog 256, rain 187, sand 243, snow 203).
Neither set is used for any choice. For CRC over $\lambda$ we use 200 random calibration/test halves per
condition, keeping near-duplicate groups together, as in the baseline runs.

% MARK: Metrics
\paragraph{Metrics} The main metric is the cost $C_\lambda$ relative to one threshold, with 95\% image-bootstrap
intervals~\cite{efron1993bootstrap}. We also report recall, precision and false alarms per image at
$\lambda \in \{1, 3, 10, 30, 100\}$; $F_\beta$ at $\beta \in \{0.5, 1, 2\}$ against no tuning and the best
in-sample constant; and calibration (expected calibration error, share real near each cut). For CRC over
$\lambda$ we report the held-out miss fraction against $\alpha$ (share of splits above), its price in false alarms
per image, and the smallest certifiable $\alpha$.

% MARK: Baselines
\paragraph{Baselines} (i) No tuning ($\tau = 0.25$, the Ultralytics default). (ii) One global CRC threshold.
(iii) The per-weather CRC threshold, Eq.~\eqref{eq:baseline-crc} (Phase-1 baseline). (iv) Option~A, the one-threshold
dial. (v) The context-only per-detection model. As reference bounds that no deployable rule can reach, we also
report the best constant threshold chosen in-sample per condition and the per-frame hindsight best.

% MARK: Ablations
\paragraph{Ablations} (a) Retraining without each feature group: class probabilities, neck components, DFL
entropy, level, context. (b) Per-condition recalibration, which targets the under-stated probabilities in DAWN fog
(17\% real at the 9.1\% cut). (c) Training-set size: 25, 50 and 100\% of frames. (d) Detector size and input
resolution: YOLOv8x at 640 versus 1280, and YOLOv8s. (e) CRC over $\lambda$ versus CRC over $\tau$ at equal
$\alpha$. Already completed: model~1 versus model~2, per-weather thresholds, and dedicated $\beta$ models.

% MARK: Compute
\paragraph{Compute} Measured on our laptop GPU (RTX 3050 Ti, 4\,GB):
\begin{itemize}
  \item YOLOv8x at 1280\,px takes 0.15--0.17\,s per frame;
  \item extracting per-box outputs for all 30.9k frames takes about 85 minutes;
  \item fitting the per-detection model takes about 11 minutes on the CPU;
  \item the CRC splits take minutes on the CPU.
\end{itemize}
We will use an NRP A100 for the new detector passes in ablation~(d) and for the repeated fits in (a)--(c).
At an estimated $\ge 10\times$ the laptop's throughput, these take a few GPU-hours in total. The laptop
remains a fallback for every step.

%% file: sections/experimental_setup.tex
% MARK: Experimental Setup and Baseline Reproduction
\section{Experimental Setup and Baseline Reproduction}\label{sec:repro}

Before building the dial we reproduced its baseline, SeqCRC~\cite{andeol2025seqcrc}, on its own benchmark: a
COCO-pretrained YOLOv8x~\cite{ultralytics2023yolov8} calibrated on MS-COCO~\cite{lin2014coco}, with the authors'
official code (\texttt{cods}, \url{https://github.com/leoandeol/cods}, commit \texttt{269ff11}). SeqCRC trains
nothing: its ``training'' is the calibration of three scalars on 2,500 images, and evaluation is on the other
2,500. Our repository, with the run scripts, logs, result files and a patch to \texttt{cods}, is
\url{https://github.com/Nicholas55555/seqcrc-baseline}.

% MARK: Setup
\subsection{Setup}\label{sec:repro-setup}
\paragraph{Hardware} A local laptop instead of a hosted GPU: an NVIDIA GeForce RTX 3050 Ti Laptop GPU (4\,GB,
driver 566.07, CUDA 12.7), an AMD Ryzen 7 4800H (8 cores), 32\,GB RAM and Windows 11. Detection ran on the GPU
(a one-off 390\,s pass); calibration ran on the CPU. The weather experiments of Section~\ref{sec:method}
use the same laptop (Section~\ref{sec:method-plan}).

\paragraph{Software} Python 3.11.15 (uv), PyTorch 2.14.1 and torchvision 0.29.1 (CUDA 12.6 wheels), Ultralytics
8.4.177, NumPy 2.4.6, numba 0.68.0, pycocotools 2.0.11 and scikit-learn 1.9.1; \texttt{cods} 0.4 at commit
\texttt{269ff11} plus our patch. The repository pins every version in \texttt{requirements.txt}.

\paragraph{Data} The full COCO val2017 set (5,000 images, SHA-256-checked): the first 2,500 images calibrate and
the last 2,500 test, which is the official code's unshuffled 50/50 split. A 300/300 subset (first 300 for
calibration, next 300 for test) served for fast checks.

\paragraph{Protocol} As in the paper (its Sec.~V-A): the official \texttt{yolov8x.pt} at Ultralytics' default
640\,px (the paper states neither), NMS IoU 0.5, confidence floor $10^{-3}$, total budget $\alpha_{\text{tot}} = 0.1$
split as $(\alpha_{\text{cnf}}, \alpha_{\text{loc}}, \alpha_{\text{cls}}) = (0.02, 0.05, 0.05)$, and mix-matching
weight 0.25. The paper's appendix (Table~VII) reports 92 YOLOv8x configurations at this budget. We ran four of them
in full (F1--F4), chosen to cover both confidence losses, the Hausdorff, LAC and Mix matchings, pixel-wise and
box-wise localization losses, additive and multiplicative boxes, and APS and LAC class sets.

% MARK: Results
\subsection{Reproduction Results}\label{sec:repro-results}

\begin{table*}[!t]
\centering\footnotesize
\setlength{\tabcolsep}{4pt}
\caption{SeqCRC with YOLOv8x on COCO val2017, $\alpha_{\text{tot}} = 0.1$: set size and test risk per task
(confidence: boxes per image, budget 0.02; localization: box stretch, budget 0.05; classification: classes per
set, budget 0.05) and global risk (budget 0.1). Paper: its appendix Table~VII. Full: the same 2,500/2,500 split as
the paper's code. 300/300: first 300 images for calibration, next 300 for test.}
\label{tab:seqcrc}
\begin{tabular}{@{}l l l rr rr rr r@{}}
\toprule
 & Matching / conf.\ loss / loc.\ loss-set / class set & Source & \multicolumn{2}{c}{Confidence}
 & \multicolumn{2}{c}{Localization} & \multicolumn{2}{c}{Classification} & Global \\
\cmidrule(lr){4-5}\cmidrule(lr){6-7}\cmidrule(lr){8-9}
 & & & size & risk & size & risk & size & risk & risk \\
\midrule
\multirow{3}{*}{F1} & \multirow{3}{*}{Hausdorff / bc-recall / pixel-add / APS} & Paper & 11.71 & 0.019 & 3.87 & 0.049 & 41.73 & 0.043 & 0.082 \\
 & & Ours, full & 11.53 & 0.020 & 1.30 & 0.053 & 40.64 & 0.042 & 0.085 \\
 & & Ours, 300/300 & 11.11 & 0.025 & 1.30 & 0.040 & 42.67 & 0.033 & 0.066 \\
\addlinespace[3pt]
\multirow{3}{*}{F2} & \multirow{3}{*}{LAC / bc-thresh / pixel-add / LAC} & Paper & 18.86 & 0.012 & 20.64 & 0.047 & 0.72 & 0.049 & 0.086 \\
 & & Ours, full & 18.02 & 0.015 & 17.69 & 0.047 & 0.75 & 0.046 & 0.083 \\
 & & Ours, 300/300 & 14.29 & 0.047 & 13.56 & 0.046 & 1.02 & 0.045 & 0.080 \\
\addlinespace[3pt]
\multirow{3}{*}{F3} & \multirow{3}{*}{Mix / bc-recall / pixel-mult / LAC} & Paper & 11.71 & 0.019 & 4.54 & 0.049 & 67.58 & 0.053 & 0.098 \\
 & & Ours, full & 11.53 & 0.020 & 1.24 & 0.052 & 66.82 & 0.055 & 0.100 \\
 & & Ours, 300/300 & 11.11 & 0.025 & 1.24 & 0.038 & 66.06 & 0.061 & 0.093 \\
\addlinespace[3pt]
\multirow{3}{*}{F4} & \multirow{3}{*}{Mix / bc-thresh / box-add / APS} & Paper & 18.86 & 0.012 & 6.12 & 0.043 & 40.75 & 0.042 & 0.078 \\
 & & Ours, full & 18.02 & 0.015 & 2.56 & 0.053 & 38.28 & 0.043 & 0.088 \\
 & & Ours, 300/300 & 14.29 & 0.047 & 2.21 & 0.055 & 40.66 & 0.038 & 0.085 \\
\bottomrule
\end{tabular}
\end{table*}

\begin{table}[!t]
\centering\footnotesize
\caption{Detector check: COCO box AP of YOLOv8x on all 5,000 val2017 images (pycocotools).}
\label{tab:det}
\begin{tabular}{@{}lrrr@{}}
\toprule
Boxes scored & AP & AP50 & AP75 \\
\midrule
Ultralytics model table (paper's detector) & 53.9 & -- & -- \\
Ours: Ultralytics predict, NMS IoU 0.7 per class & 53.2 & 69.8 & 58.0 \\
Ours: Ultralytics predict, class-agnostic NMS 0.5 & 51.6 & 68.8 & 55.6 \\
Ours: \texttt{cods} wrapper, batch 1 (SeqCRC input) & 45.1 & 65.9 & 48.2 \\
Ours: \texttt{cods} wrapper, batch 8 & 10.7 & 32.1 & 1.4 \\
\bottomrule
\end{tabular}
\end{table}

Table~\ref{tab:seqcrc} compares our runs with the paper. On the full 2,500/2,500 split the confidence and
classification steps reproduce closely: the box-count-recall step keeps 11.53 boxes per image in F1 and F3 (paper
11.71) and the box-count-threshold step 18.02 in F2 and F4 (paper 18.86), and the class sets hold 40.6, 0.75, 66.8
and 38.3 classes in F1--F4 (paper 41.7, 0.72, 67.6 and 40.8). Test risks sit at or near their budgets, as in the
paper: global risks are 0.085, 0.083, 0.0997 and 0.088 against the 0.1 budget (paper 0.082, 0.086, 0.098 and
0.078), and a few single-split task risks are slightly above theirs (at most 0.055 against 0.05), as is the paper's
own F3 classification risk (0.053). Localization is the exception: our stretches are smaller in every
configuration (F1: 1.30 against 3.87; F3: 1.24 against 4.54; F4: 2.56 against 6.12), which
Section~\ref{sec:repro-disc} traces to the batch size.

% MARK: Discrepancies
\subsection{Discrepancies}\label{sec:repro-disc}
\paragraph{Localization sets} Our localization sets are much smaller than reported for YOLOv8x, while the class
sets and box-count-recall confidence sets are close. The likely cause is in the official YOLO wrapper
(\texttt{cods/od/models/yolo.py}): it rescales the raw head output by the ratio of original to letterboxed image
size and never removes the letterbox padding. With batch size 1 (our runs) Ultralytics pads each image only to a
multiple of 32, so the error is small; with a batch of differently sized images it pads every image to
$640\times640$ and the boxes land far from the objects. Table~\ref{tab:det} shows the effect directly: the same
wrapper scores 45.1 AP at batch 1 but 10.7 AP (AP75 1.4) at batch 8. Re-running F3 on the 300/300 subset with batch
8 raises the stretch from 1.24 to 4.70, close to the paper's 4.54, while the confidence and class sets barely move
(11.11 vs.\ 11.46 boxes, 66.1 vs.\ 65.7 classes). The paper's best YOLOv8x stretches (3.87--5.45, its Table~III)
are also 3--4 times those of DETR-101 (1.04--1.55), whose wrapper rescales normalized boxes and is unaffected. The
paper does not state its batch size, but the official tutorial notebook calls \texttt{build\_predictions} with
\texttt{batch\_size=12}, so its YOLOv8x localization numbers were most likely produced with a batch size above 1.
This rests on one configuration on 300/300 images; a full-size batch-8 run did not finish in time.

\paragraph{The detector itself} Through Ultralytics we measure 53.2 AP against the published 53.9
(Table~\ref{tab:det}; we use \texttt{predict()} with per-image letterboxing and pycocotools, not \texttt{val()}).
The boxes SeqCRC actually receives score 45.1: class-agnostic NMS at IoU 0.5 (the paper's IoU; the \texttt{cods}
NMS is class-agnostic) costs 1.6 AP, and the wrapper's box rescaling (padding not removed, separate $x$ and $y$
scales, no clipping) costs the remaining 6.5. Its scores and classes are the same maximum sigmoid score and argmax
that Ultralytics uses. We kept the official wrapper because the paper's numbers come from it.

\paragraph{One split} The paper reports a single calibration/test draw, as we do. The guarantee holds in
expectation over draws, so single-split risks may sit slightly above or below the budget; on 300 test images the
fluctuations are larger (e.g., box-count-threshold risk 0.047 against 0.02).

\paragraph{Software drift} We ran the code at commit \texttt{269ff11} (May 22, 2026; library code last changed Oct.
22, 2025) with Ultralytics 8.4.177; the paper's runs used earlier versions, which it does not state. Our two speed
patches changed no output: the mask selection is exact, and the float64 running sum moved calibration risks by
about $2\times10^{-9}$, leaving every output of a 300/300 check identical (Appendix~\ref{app:repro}).

%% file: sections/appendix_reproduction.tex
% Phase 1: reproduction issues, fixes and evidence (Appendix A of the paper, and of phase1_baseline.tex).
% Must follow \appendices. Drop it from main.tex for the final 6-page paper; the repository keeps the details.

% MARK: Appendix: Reproduction Issues and Fixes
\section{Reproduction Issues and Fixes}\label{app:repro}
These are the problems we met while reproducing SeqCRC (Section~\ref{sec:repro}) and how we solved them. The
patch, scripts and logs are in our repository.
\begin{enumerate}
\item \textbf{Hard-coded path and logging.} The official evaluation script, \path{cods/od/evaluate.py}, reads
COCO from \path{/datasets/shared_datasets/coco/} and requires Weights \& Biases. \emph{Fix:} our
\path{scripts/run_seqcrc.py} repeats its \texttt{run\_experiment} step for step with a \texttt{-{}-coco} argument
and JSON-lines output, and adds the paper's $10^{-3}$ confidence floor, which \path{evaluate.py} never applies (the
tutorial notebook uses $4\times10^{-2}$).
\item \textbf{Calibration crashes at the current commit} with a \texttt{TypeError} (``expected Tensor \dots\ but
got list''). Lint commit \texttt{e70d773} (Oct.\ 20, 2025) turned \texttt{x = list([\dots],)} into
\texttt{x = ([\dots],)} in \texttt{build\_predictions}, wrapping every per-image list in a 1-tuple; the tutorial
notebook only works from older cached predictions. \emph{Fix:} unwrap the tuples when \texttt{ODPredictions} is
built (no upstream edit).
\item \textbf{4\,GB GPU and device mismatch.} Unfiltered YOLOv8x outputs (hundreds of boxes per image, each with
a 91-way class vector) do not fit, and ground truth and boxes ended up on different devices. \emph{Fix:} apply the
paper's $10^{-3}$ floor right after NMS (the same box set), keep boxes on the CPU, then move all predictions to one
device; this leaves 50.8 boxes per test image on average (median 30, maximum 741).
\item \textbf{Calibration too slow for a laptop.} A 50/50-image smoke test was still calibrating after about
12\,min on the GPU and was stopped. Profiling with \texttt{py-spy} pointed to (a) a Python loop over every box for
every candidate threshold and (b) a full re-summation of all $n$ per-image losses after each single update,
$O(n^2)$ overall (100 images: 294\,s; 300: 1,151\,s; extrapolated to 2,500: about 8\,h per configuration).
\emph{Fix} (\path{patches/cods-reproduction.patch}): boolean-mask indexing, which selects the same rows in the same
order, and a running sum of the losses. Unit tests show identical selections and sums (within $10^{-6}$), and a
300/300 run with and without the running sum gives the same step-1 thresholds to every printed digit. Calibration
then runs on the CPU, one thread per process. The four full runs ran in parallel, with diagnostic runs alongside,
so their wall-clock times (143--253\,min) were measured under CPU contention.
\item \textbf{Smaller snags.} With 10 calibration images \texttt{cods} reports ``No good lambda found'', correctly:
SeqCRC needs $\alpha_{\text{loc}} \ge \alpha_{\text{cnf}} + 1/(n+1)$, so smoke tests use $n \ge 100$. On Windows,
\texttt{python} was the Microsoft Store stub and console output failed on \texttt{cp1252}; we used a uv-managed
CPython and \texttt{PYTHONIOENCODING=utf-8}.
\end{enumerate}

\begin{figure}[!t]
\centering
\includegraphics[width=\columnwidth]{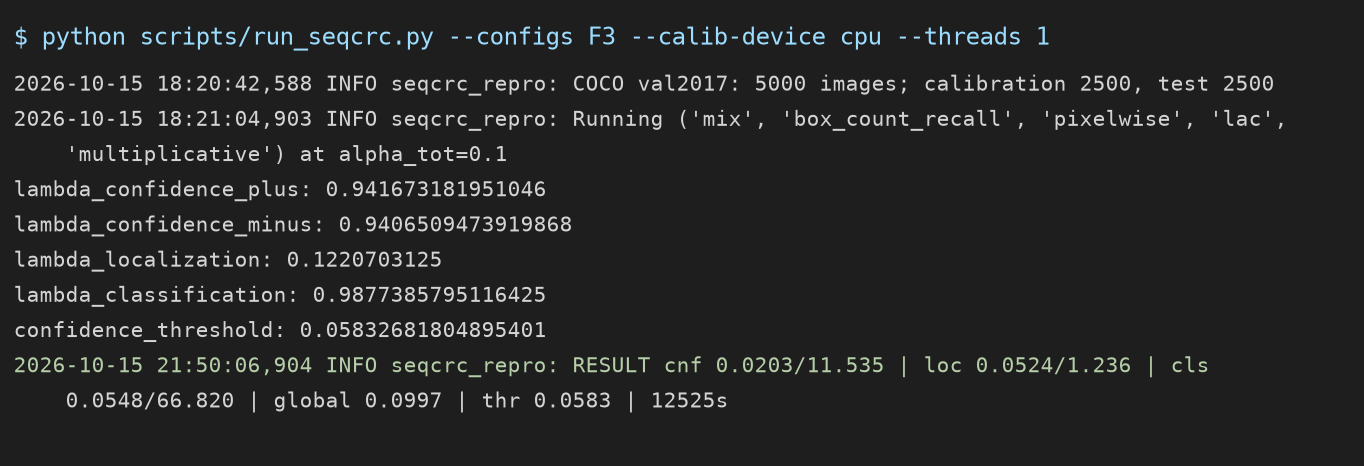}\\[4pt]
\includegraphics[width=\columnwidth]{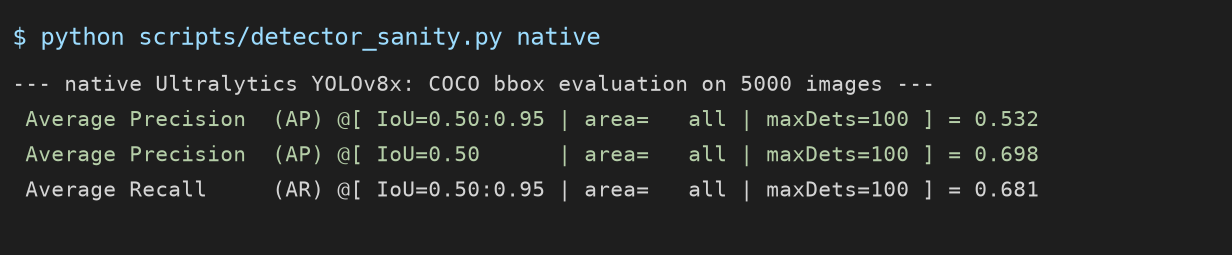}
\caption{Evidence: excerpts of the saved logs (lines selected and wrapped, text unchanged; the first line names the
command). Top: a full-size SeqCRC run, F3 (\texttt{logs/10\_full\_F3.log}). Bottom: COCO evaluation of YOLOv8x
through Ultralytics (\texttt{logs/11\_detector\_sanity\_native.log}).}
\label{fig:evidence}
\end{figure}

Fig.~\ref{fig:evidence} shows excerpts of the saved logs for one full-size SeqCRC run and for the detector check.

